\documentclass[conference]{IEEEtran}
\usepackage{cite}
\usepackage{graphicx}
\usepackage{amsmath}
\usepackage{amsmath}
\usepackage{amsfonts}
\usepackage{amssymb}
\usepackage{multirow}
\usepackage{float}
\usepackage[linesnumbered,ruled,vlined]{algorithm2e}
\usepackage{caption}
\usepackage{url}
\begin{document}
\bstctlcite{IEEEexample:BSTcontrol}
\RestyleAlgo{ruled} 
\title{A Comparative Evaluation of FedAvg \\ and Per-FedAvg Algorithms for Dirichlet Distributed Heterogeneous Data}

\author{
\hspace{-0.75cm}
Hamza Reguieg$^1$, Mohammed El Hanjri$^2$, Mohamed El Kamili$^1$, Abdellatif Kobbane$^2$\\
$^1$Higher School of Technology, Hassan II University in Casablanca, Morocco \\
$^2$ENSIAS, Mohammed V University in Rabat, Morocco.\\
\vspace{0.1cm}
\textit {hamza.reguieg-etu@etu.univh2c.ma, mohammed.elhanjri@um5r.ac.ma,}\\
\textit{abdellatif.kobbane@ensias.um5.ac.ma, mohamed.elkamili@univh2c.ma}
}

\maketitle

\begin{abstract}
In this paper, we investigate Federated Learning (FL), a paradigm of machine learning that allows for decentralized model training on devices without sharing raw data, thereby preserving data privacy. In particular, we compare two strategies within this paradigm: Federated Averaging (FedAvg) and Personalized Federated Averaging (Per-FedAvg), focusing on their performance with Non-Identically and Independently Distributed (Non-IID) data. Our analysis shows that the level of data heterogeneity, modeled using a Dirichlet distribution, significantly affects the performance of both strategies, with Per-FedAvg showing superior robustness in conditions of high heterogeneity. Our results provide insights into the development of more effective and efficient machine learning strategies in a decentralized setting.
\end{abstract}

\IEEEpeerreviewmaketitle

\section{Introduction}
Machine learning (ML), a subset of artificial intelligence, has revolutionized numerous fields, from healthcare to finance, by enabling computers to learn from data and make intelligent decisions or predictions \cite{hullermeier2019, sideygibbons2019}. Traditional ML approaches typically rely on centralized models trained on large, aggregated datasets. However, these centralized approaches often face significant challenges related to data privacy, security, and data ownership \cite{sideygibbons2019}. Moreover, the uncertainty inherent in machine learning models, both aleatoric and epistemic, has been a topic of considerable research \cite{hullermeier2019}.\\

To address these challenges, a novel paradigm in machine learning, known as Federated Learning (FL), has emerged. FL allows for model training on decentralized data residing on local devices, such as mobile phones or IoT devices, without the need to share raw data \cite{niknam2019}. This approach not only preserves data privacy but also enables the utilization of rich, diverse data sources that would otherwise be inaccessible due to privacy concerns or data transfer limitations \cite{niknam2019}.\\

In FL, a global model is collaboratively trained across multiple devices or 'nodes', each holding their local data. The nodes compute model updates locally and only share these updates with a central server, where they are aggregated to update the global model \cite{niknam2019}. This process is iteratively performed until the model's performance converges. As such, FL effectively addresses the privacy-security-utility trade-off, making it an attractive approach for applications dealing with sensitive data, such as healthcare or finance \cite{niknam2019, hanjri2023federated, el2023new}.\\

However, FL is not without its challenges. Issues such as system heterogeneity, communication efficiency, and straggler mitigation need to be addressed for effective FL deployment \cite{niknam2019}. Furthermore, despite the privacy advantages of FL, potential security threats, such as data poisoning and adversarial attacks, still exist and require further research \cite{lyu2020, kabbajdistfl}. In response to these challenges, researchers have proposed secure aggregation protocols for privacy-preserving machine learning, which are communication-efficient and failure-robust \cite{bonawitz2017}.\\

One of the key strategies within the FL framework is Federated Averaging (FedAvg), introduced by McMahan et al. \cite{bonawitz2017}. FedAvg is a robust method that combines local updates computed through stochastic gradient descent (SGD) on each client with a server that performs model averaging. This algorithm is particularly effective in scenarios where data is unbalanced and non-IID (Independent and Identically Distributed), and it can significantly reduce the rounds of communication needed to train a deep network on decentralized data \cite{bonawitz2017}.\\

In this work, we meticulously compare the FedAvg strategy with another aggregation technique, specifically examining their suitability for Non-IID data with Dirichlet distribution. Through this comparison, we aim to shed light on the strengths and weaknesses of each strategy, providing valuable insights for the development of more effective and efficient machine learning strategies in a decentralized setting.\\

The rest of this paper is organized as follows, with section II discussing related works with a focus on various Federated Learning Algorithms. The system model is described in Section III, while Section IV presents the simulation and numerical findings. Section V serves as the paper's conclusion.

\section{Related Works}
Federated Learning (FL) is a machine learning paradigm that has seen significant advancements over the years. This approach allows for decentralized learning, where the data remains on the local devices, and only model updates are shared. This approach has gained significant attention due to its ability to maintain privacy and reduce communication overhead. Various aggregation algorithms have been proposed in the literature to improve the performance and efficiency of FL.\\

The journey of these advancements can be traced back to 2017, with the introduction of the FedAvg algorithm in \cite{bonawitz2017}. This algorithm, which averages the model updates from each client and applies the average update to the global model, sets the foundation for many subsequent FL algorithms. It was a significant step forward in the development of FL algorithms, providing a simple yet effective method for aggregating model updates in a decentralized setting.\\

However, the FedAvg algorithm assumes that all clients have identically distributed data, which is often not the case in real-world scenarios. Recognizing this limitation, the FedAvgM algorithm was proposed in 2019 in \cite{fedavgm}. This algorithm introduced a momentum term to stabilize the learning process in the presence of non-identical data distribution. This modification to the FedAvg algorithm marked a significant step forward in the development of FL algorithms, addressing one of the key challenges in FL.\\

In 2020, the FedProx algorithm was introduced in \cite{fedprox}. This algorithm further improved upon FedAvg by introducing a proximal term to the optimization objective. This term encourages the local updates to stay close to the global model, thereby improving the robustness of the algorithm in heterogeneous networks. The introduction of the FedProx algorithm was a significant advancement in FL, addressing the challenge of network heterogeneity, which is common in real-world FL scenarios. The same year, the SCAFFOLD algorithm was introduced in \cite{scaffold}. This algorithm used control variates to reduce the variance of the updates, leading to faster convergence and improved performance. It was particularly effective in scenarios with high client drop-out rates, a common issue in FL. The introduction of the SCAFFOLD algorithm marked another significant milestone in the development of FL algorithms, addressing the challenge of high client drop-out rates.\\

In 2021, the MOON algorithm was presented in \cite{moon}. This algorithm introduced a contrastive loss function that encouraged the global model to learn from the differences between the local models. This approach showed promising results in visual classification tasks, demonstrating the potential of FL in complex machine learning tasks. The introduction of the MOON algorithm marked a significant advancement in FL, expanding its applicability to more complex tasks. Later in 2021, the FedDyn algorithm was introduced in \cite{fedyn}. This algorithm used dynamic regularization to adapt the learning process based on the heterogeneity of the data. This approach improved the performance of FL in scenarios with skewed data distribution, a common issue in real-world FL scenarios. The introduction of the FedDyn algorithm marked another significant milestone in the development of FL algorithms, addressing the challenge of skewed data distribution.

In the same year, the FedGen algorithm was introduced in \cite{fedgen}. This algorithm used generative models to provide strong privacy guarantees in FL, allowing clients to share synthetic data instead of real data, thereby preserving the privacy of the clients' data. The introduction of the FedGen algorithm marked a significant advancement in FL, addressing the critical challenge of privacy preservation.\\

In 2022, the FedLC algorithm was presented in \cite{fedlc}. This algorithm addressed the issue of label distribution skew in FL by introducing a logit calibration mechanism that adjusted the local models based on the global label distribution, thereby improving the fairness of the learning process. The introduction of the FedLC algorithm marked another significant milestone in the development of FL algorithms, addressing the challenge of label distribution skew.\\

On the personalized FL front, the Per-FedAvg algorithm was introduced in \cite{fedavg}. This algorithm proposed a model-agnostic meta-learning approach to personalize federated learning, providing theoretical guarantees for its performance. The introduction of the Per-FedAvg algorithm marked a significant shift in the FL paradigm, moving towards personalized FL.\\

The evolution of FL has been marked by continuous innovation, with each new algorithm building on the strengths of its predecessors while addressing their limitations. This paper focuses on comparing the FedAvg and Per-FedAvg algorithms for non-IID data. FedAvg laid the groundwork for FL, while Per-FedAvg introduced the concept of personalized FL, marking a significant shift in the FL paradigm. These algorithms, along with the others mentioned, have significantly expanded the applicability and effectiveness of FL.

\section{System Model}
In this section, we will elaborate on the system models that form the foundation of our comparison. Our primary focus is the Federated Averaging (FedAvg) approach. We discuss its limitations, particularly under the condition of high data heterogeneity, and introduce the Personalized Federated Averaging (Per-FedAvg) as an alternative that addresses these shortcomings. We will also discuss the different data distribution models to evaluate and compare the performance of FedAvg and Per-FedAvg, including one based on a Dirichlet distribution. Detailed explanations of these approaches and their implementation follow
\subsection{The FedAvg Approach}

This approach presumes a synchronized update procedure that transpires in communication rounds. It involves a static pool of $K$ clients, each possessing a fixed local dataset. At the initiation of every round, a random proportion $C$ of clients is chosen, and the server disseminates the current global algorithm state to them, such as the prevailing model parameters. We elect to involve only a fraction of clients for improved efficiency, as our tests exhibit a saturation point beyond which the inclusion of additional clients provides no added benefits. Each chosen client then undertakes local computations based on the global state and its local dataset and transmits an update to the server. The server consequently applies these updates to its global state, and the cycle continues. Although our main emphasis is on non-convex neural network goals, the algorithm under review can be applied to any finite-sum objective of the following form:
\begin{equation}
\min _{w \in \mathbb{R}^{d}} f(w) \quad \text { where } \quad f(w) \stackrel{\text { def }}{=} \frac{1}{n} \sum_{i=1}^{n} f_{i}(w)
\label{1}
\end{equation}
For a machine learning issue, we usually define $f_{i}(w)=$ $\ell\left(x_{i}, y_{i} ; w\right)$, that is, the loss of the prediction on example $\left(x_{i}, y_{i}\right)$ made with model parameters $w$. We presume that there are $K$ clients over which the data is divided, with $\mathcal{P}{k}$ representing the set of indexes of data points on client $k$, with $n{k}=\left|\mathcal{P}_{k}\right|$. Consequently, we can rephrase the objective (\ref{1}) as 
\begin{equation}
f(w)=\sum_{k=1}^{K} \frac{n_{k}}{n} F_{k}(w) 
\end{equation}

where $$\quad F_{k}(w)=\frac{1}{n_{k}} \sum_{i \in \mathcal{P}_{k}} f_{i}(w)$$

When the partition $\mathcal{P}_{k}$ was created by randomly distributing the training examples across the clients, then we would have $\mathbb{E}_{\mathcal{P}_{k}}\left[F_{k}(w)\right]=f(w)$, where the expectation is over the set of examples assigned to a fixed client $k$. This IID assumption is commonly made by distributed optimization algorithms; we refer to situations where this assumption is violated (i.e., $F_{k}$ could be a significantly poor approximation to $f$ ) as the non-IID setting.
\subsection{The Per-FedAvg Approach}
The basic concept underpinning the Model-Agnostic Meta-Learning (MAML) approach, as presented in \cite{maml}, can be leveraged to create a personalized version of the FL challenge. First, let's review the MAML model. MAML, unlike the standard supervised learning approach, aims to find an initial model that adapts quickly to a new task within a fixed computational resource, using tasks taken from a common distribution. Instead of a model that works well across all tasks, the focus in MAML is to find an initialization that's effective post-update when a new task is introduced.\\

The formula used is: 
\begin{equation}
\min _{w \in \mathbb{R}^{d}} F(w):=\frac{1}{n} \sum_{i=1}^{n} f_{i}\left(w-\alpha^{'} \nabla f_{i}(w)\right),
\label{3}
\end{equation}

Here, the stepsize is represented as $\alpha^{'} \geq 0$. This equation's strength lies in the fact that it retains the benefits of FL and recognizes the variation among users. Users can use the solution of this new problem as a starting point and fine-tune it using their own data.\\

Solving the original problem (\ref{1}) for the considered heterogeneous data model isn't the ideal choice as it doesn't quickly adapt to each user's local data, even after a few steps of the local gradient. However, solving (\ref{3}) provides an initial model (meta-model) that can be quickly adapted to each user after one step of the local gradient. The issue can also be extended for users who run multiple steps of gradient updates, but for simplicity, we focus on the single gradient update scenario.



Subsequently, a novel algorithm titled Personalized FedAvg (Per-FedAvg) aimed at resolving equation (\ref{3}). This is an enhanced version of the conventional FedAvg, particularly engineered to ascertain the solution to equation (\ref{3}) as opposed to equation (\ref{1}). Mirroring the FedAvg process, the server chooses a user subset in each cycle, dispatches the existing model to these users who then tweak the model based on their unique loss function by carrying out steps of stochastic gradient descent. These modified models are relayed back to the server, which then refines the global model by averaging all the received models. This entire sequence is then iterated.
\begin{algorithm}
\caption{The Per-FedAvg Algorithm}
Initial iterate $w_0$, fraction of active users $r$.\\
\For{$k: 0$ to $K-1$}{
The server randomly selects a subset of users, denoted as $\mathcal{A}_k$, where the size of the subset is $r n$;
Server sends $w_k$ to all users in $\mathcal{A}_k$;\\
\For{all $i \in \mathcal{A}_k$}{
Set $w_{k+1,0}^i=w_k$;\\
\For{$t: 1$ to $\tau$}{
Compute the stochastic gradient $\tilde{\nabla} f_i\left(w_{k+1, t-1}^i, \mathcal{D}_t^i\right)$ using dataset $\mathcal{D}_t^i$;\\
Set $\tilde{w}_{k+1, t}^i=w_{k+1, t-1}^i-\alpha^{'} \tilde{\nabla} f_i\left(w_{k+1, t-1}^i, \mathcal{D}_t^i\right)$;\\
Set $w_{k+1, t}^i=w_{k+1, t-1}^i-\beta\left(I-\alpha^{'} \tilde{\nabla}^2 f_i\left(w_{k+1, t-1}^i, \mathcal{D}_t^{\prime \prime} i\right)\right)
\times \tilde{\nabla} f_i\left(\tilde{w}_{k+1, t}^i, \mathcal{D}_t^{\prime}\right)$;
}
User $i$ sends $w_{k+1, \tau}^i$ back to server;
}
Server updates its model by calculating the mean of the received models: $w_{k+1}=\frac{1}{r n} \sum_{i \in \mathcal{A}_k} w_{k+1, \tau}^i$;
}
\end{algorithm}


In \cite{fedavg}, the authors altered the data distribution for half of the users to augment data heterogeneity. Precisely, users possessing $a / 2$ images from any of the initial five classes had these images eliminated from their dataset. This alteration proved that under these novel distributions, Per-FedAvg outperforms FedAvg, yielding a more individualized solution. Nevertheless, the imposed data distribution structure may not fully represent the intricacy and unpredictability of real-world situations.
\begin{figure}[h]
    \center
    \includegraphics[width=0.5\textwidth]{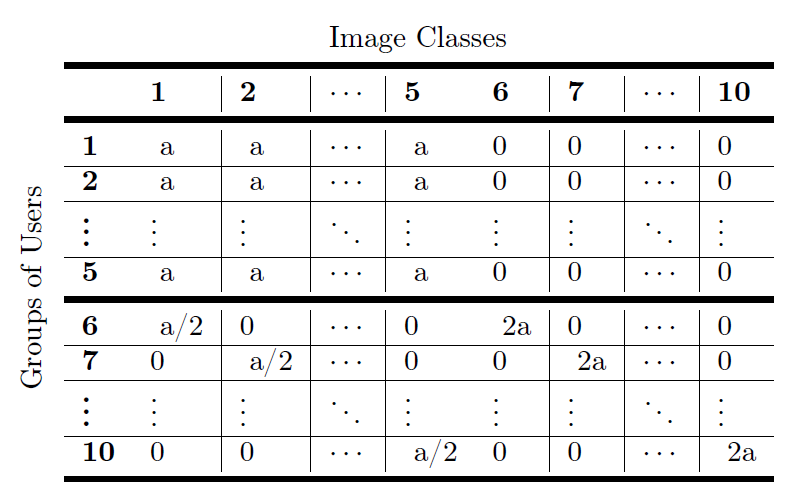}
    \caption{Illustration of the experiment’s setting in \cite{fedavg}}
\end{figure}

To address this, we use the approach based on \cite{dirichlet}: data are selected randomly based on a Dirichlet distribution. This approach is derived from a previous study where class labels were assumed to be drawn independently following a categorical distribution over $N$ classes, parameterized by a vector $q$. This vector is subject to the constraints $q_i \geq 0$ and $\left\lVert q \right\rVert_1  = 1$. It is then drawn from a Dirichlet distribution $q \sim $ Dir$(\alpha p)$, which $p$ is defined by a prior class distribution over $N$ classes and a concentration parameter $\alpha$ that controls the similarity among clients.

In this model, for every client, we determine a concentration parameter $\alpha$, and from this, we sample $q$ and assign the client with a corresponding number of data instances across multiple classes. This approach enables us to capture a wide spectrum of data heterogeneity, ranging from complete similarity (with a large concentration parameter) to extreme heterogeneity (with a small concentration parameter).
\begin{figure*}[h]
    \center
    \includegraphics[width=1\textwidth]{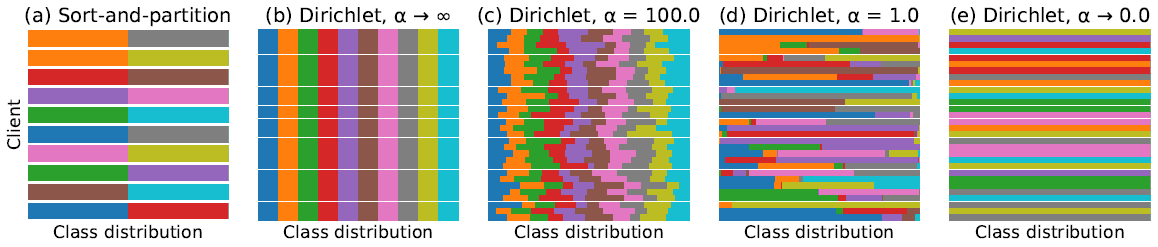}
    \caption{Populations made up of non-identical clients. Different colors are used to depict distribution amongst classes. (A) Ten customers, each assigned two classes, were produced by the sort-and-partition scheme. (b – e) Populations derived from Dirichlet distributions, each with 30 randomly chosen clients and various concentration factors.\cite{dirichlet}}
\end{figure*}

By adopting this methodology, our comparison between FedAvg and Per-FedAvg will encompass a broader and more realistic range of data distributions. This provides an accurate understanding of the relative strengths and weaknesses of these two aggregation methods under various conditions of data heterogeneity.

\section{Simulation and Numerical Results}

\subsection{Simulation Setup}
All experiments were conducted on an Intel(R) Core(TM) i7-7700HQ CPU @ 2.80GHz with hardware specifications of 4 cores and 8 threads, backed by 16GB RAM.

\subsection{Dataset}
Our study focused on a visual classification task using the CIFAR-10 dataset. The dataset, configured with an $\alpha$ parameter of $0.5$ according to a Dirichlet distribution, emulates the non-IID nature often encountered in federated learning scenarios. This dataset comprises 60,000 color images of 32x32 pixels, spread across ten classes, thereby providing a robust benchmark for our algorithm comparison.

\subsection{Algorithm Implementation}
The implementation was executed using Python 3 and TensorFlow, along with the Keras API. We utilized the LeNet-5 model, a standard convolutional neural network that is commonly applied in image classification tasks, in both FedAvg and Perf-FedAvg methods.

\subsection{Parameters and Configuration}
In our experiment, both FedAvg and Perf-FedAvg completed 1000 global epochs and 10 local epochs per round. The learning rate was set at 0.01 for local updates, and we maintained a uniform batch size of 40 across all clients in our simulated federated learning network. We selected a fraction of 0.5 of the clients to participate in each training round.

Our decision to adopt Per-FedAvg (Hessian Form) over Per-FedAvg (First Order Form) was not arbitrary but rather strategically guided. The primary factor influencing this choice was its demonstrably improved performance in environments marked by high data heterogeneity. Perf-FedAvg (Hessian Form) has an added advantage over Per-FedAvg (First Order Form) because it incorporates a second-order approximation of the gradient into its computations, using the Hessian matrix. This allows for more precise and effective model updates, especially in complex and diverse data landscapes. The Hessian matrix approximates the curvature of the loss function, making the algorithm more resilient to the variance in gradients that often accompanies highly heterogeneous data. Consequently, this leads to improved performance and stability of the learning process, justifying our preference for Perf-FedAvg (Hessian Form).







\subsection{Results and Analysis}
\begin{figure}[H]
    \centering
    \includegraphics[width=1\linewidth]{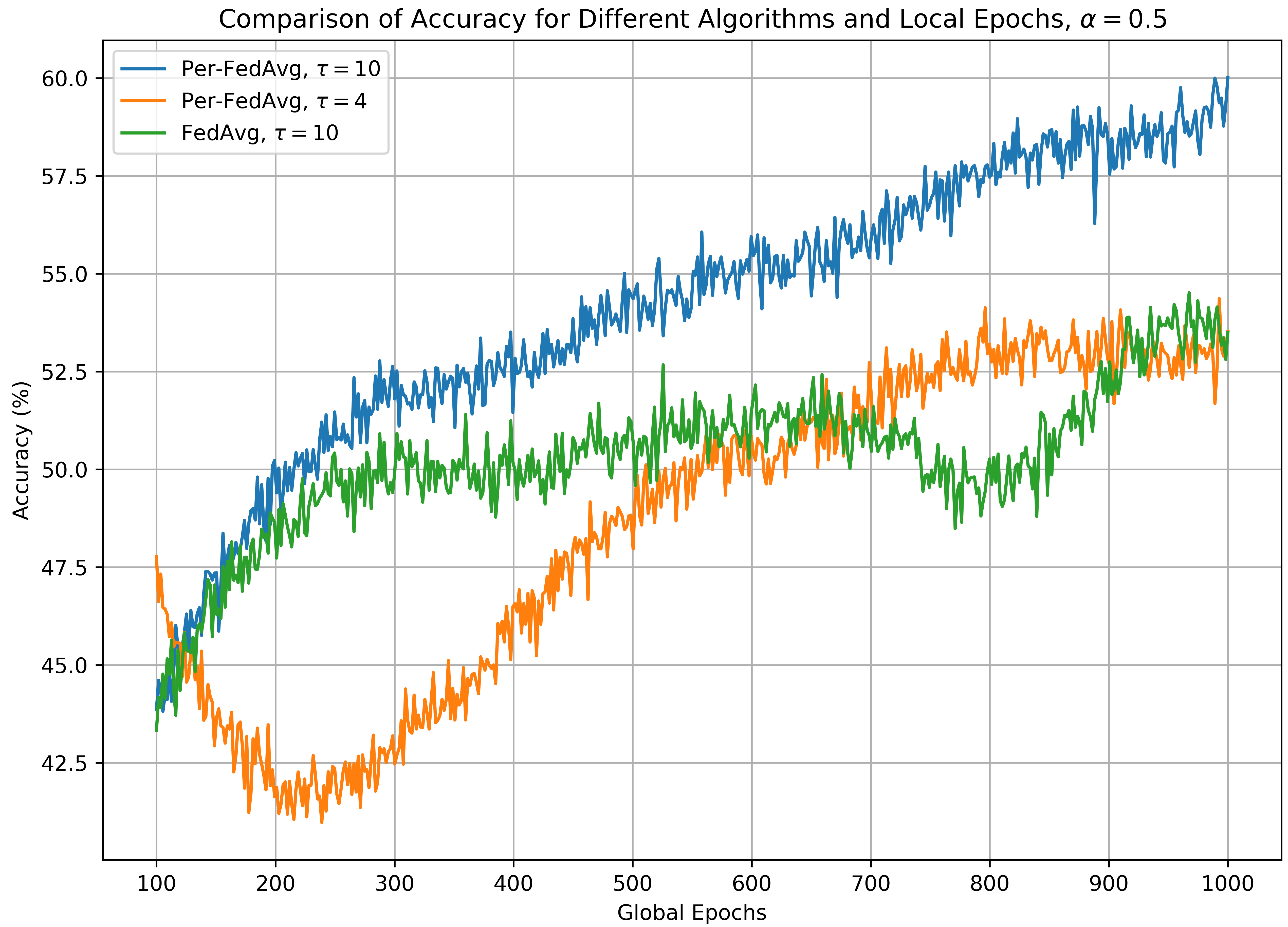}
    \caption{The FedAvg and Per-FedAvg Performance for heterogeneous data ($\alpha=0.5$), $\tau$ represents the number of local updates.}
    \label{plot05}
\end{figure}

\begin{figure}[H]
    \centering
    \includegraphics[width=1\linewidth]{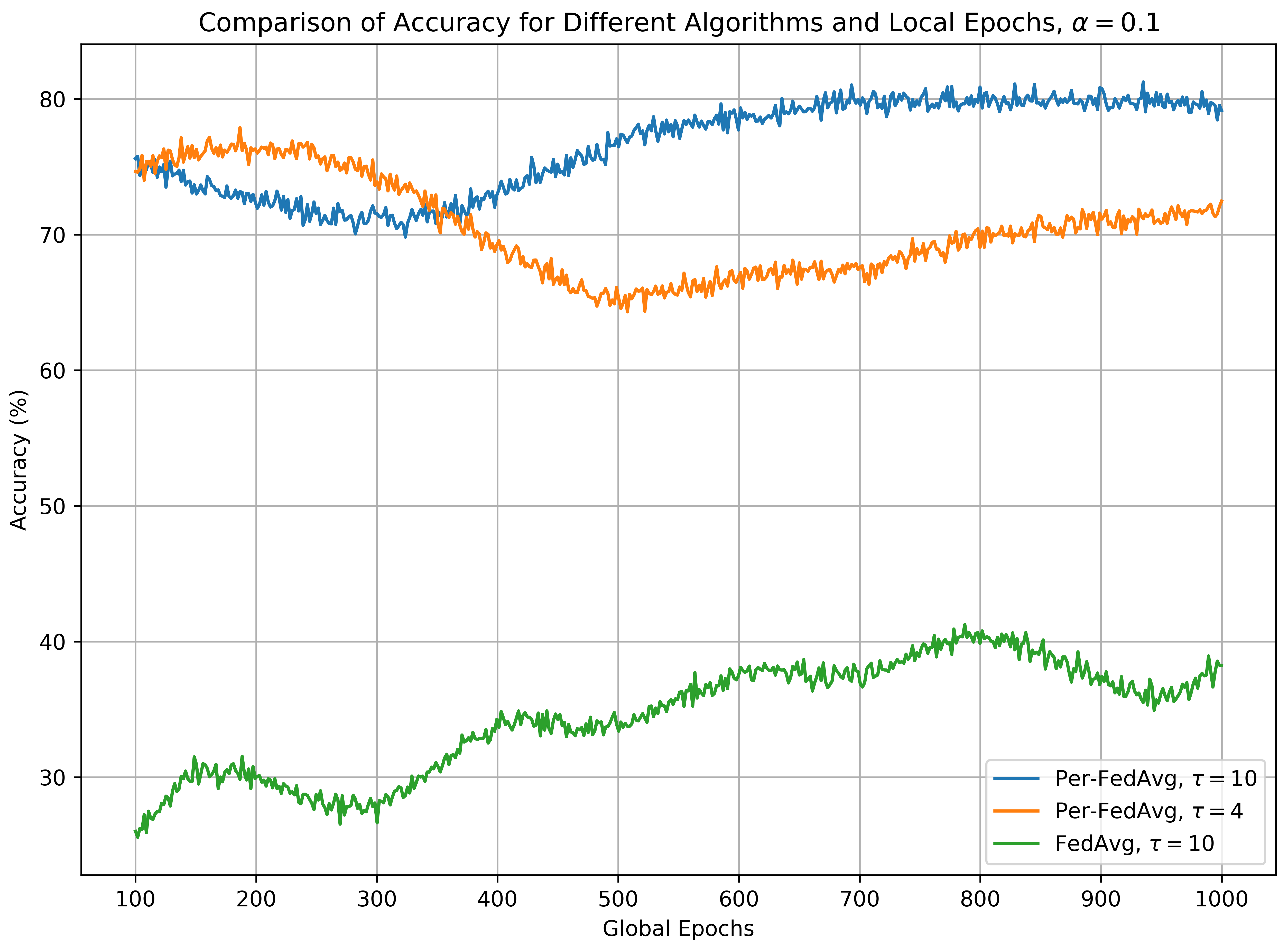}
    \caption{The FedAvg and Per-FedAvg Performance for highly heterogeneous data ($\alpha=0.1$), $\tau$ represents the number of local updates.}
    \label{plot01}
\end{figure}

Before delving into the analysis of performance under varying conditions of heterogeneity, we first consider the case of high data homogeneity represented by $\alpha=10$. In this scenario, FedAvg presented a consistent performance following numerous global epochs. Conversely, Per-FedAvg peaked at a lower accuracy, indicating that in a scenario where data points exhibit high similarity, the Per-FedAvg algorithm may not perform as optimally as FedAvg.
\begin{figure}[H]
    \centering
    \includegraphics[width=1\linewidth]{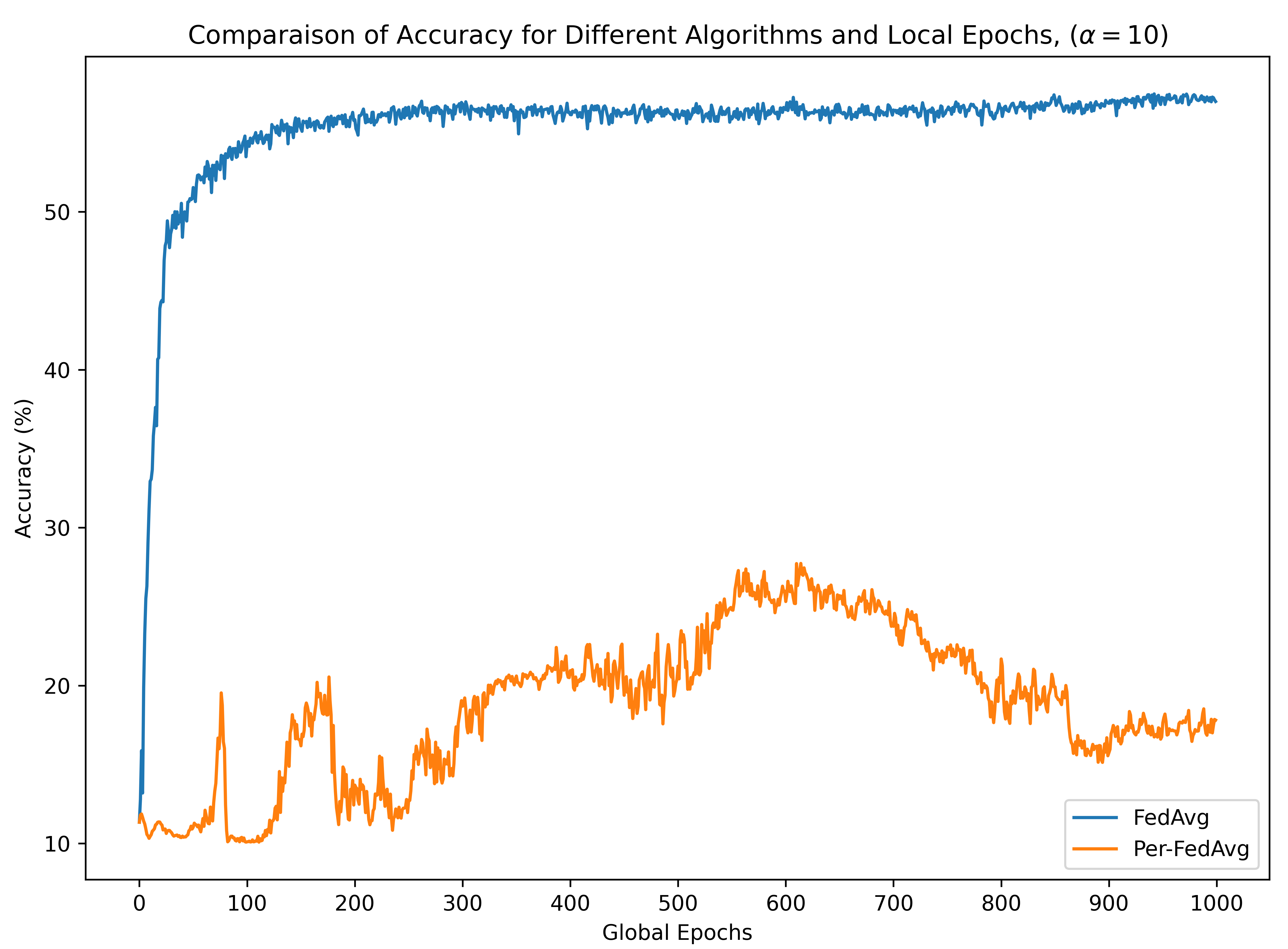}
    \caption{The FedAvg and Per-FedAvg Performance for homogeneous data ($\alpha=10$) }
    \label{plot}
\end{figure}
Building on this understanding, we now explore the scenarios of higher data heterogeneity.\\
In figure \ref{plot01}, Per-FedAvg algorithm exhibits superior performance under conditions of high heterogeneity ($\alpha=0.1$), with accuracy levels maintained around $75\%-80\%$ for $10$ local updates. Interestingly, decreasing the number of local updates to $4$ results in lower accuracy, particularly during the middle epochs, before stabilization around $72\%$. This suggests that Per-FedAvg can effectively learn from diverse data instances, but is dependent on a sufficient number of local updates to maintain high performance.\\
In contrast, when we increase $\alpha$ to $0.5$, representing a move toward data homogeneity, Per-FedAvg manifests different patterns of performance as shown in figure \ref{plot05}. For the case of $4$ local updates, the algorithm begins with higher accuracy, before settling at around $53\%$. With 10 local updates, however, the algorithm demonstrates steady improvement over the epochs, achieving almost $60\%$ accuracy by the final epoch. This indicates that, while Per-FedAvg can handle homogeneous data, the number of local updates becomes a more significant factor in performance, with greater local learning (more updates) enhancing the accuracy.
The figure \ref{plot} shows that the performance of the FedAvg algorithm, with a fixed $10$ local updates, varies notably depending on the $\alpha$ parameter. With higher heterogeneity ($\alpha=0.1$), FedAvg underperforms relative to Per-FedAvg, suggesting this algorithm may struggle with diverse data. However, the increase in $\alpha$ to $0.5$ results in substantial improvement in accuracy, with results closely aligned to the performance of Per-FedAvg with $4$ local updates.
Our experiment revealed that the level of data heterogeneity introduced by the Dirichlet distribution significantly influences the performance of both FedAvg and Perf-FedAvg. Specifically, Perf-FedAvg demonstrated superior robustness in the face of data heterogeneity, outperforming FedAvg in scenarios of high heterogeneity. This result highlights Perf-FedAvg as a promising choice for federated learning in non-IID conditions.

\begin{table}[H]
\centering
\caption{Comparison of test accuracy of different algorithms given different parameters}
\begin{tabular}{|c|c|c|}
\hline
\textbf{Parameters} & \textbf{FedAvg} & \textbf{Per-FedAvg}  \\
\hline
$\alpha=0.5$, $\tau=10$ & 53.24\% & 59.42 \% \\
$\alpha=0.5$, $\tau=4$ & - & 53.20 \%  \\
$\alpha=0.1$, $\tau=10$ & 38.31\% & 79.73 \% \\
$\alpha=0.1$, $\tau=4$ & - & 72.07 \% \\
\hline
\end{tabular}

\end{table}

\section{Conclusion}
Through our comparison of the FedAvg and Per-FedAvg strategies, we find that the level of data heterogeneity, represented by the $\alpha$ parameter in the Dirichlet distribution, significantly impacts the performance of both strategies. Particularly, Per-FedAvg outperforms FedAvg under conditions of high heterogeneity. Our findings indicate that Per-FedAvg exhibits superior performance in diverse data environments, making it a promising choice for FL in non-IID conditions. However, we also observe that the number of local updates becomes a significant factor in the performance of Per-FedAvg when handling heterogeneous data, indicating areas for further investigation and optimization. Overall, our work provides valuable insights for future research in developing more effective FL strategies, taking into account data heterogeneity and the dynamics of local updates.

\section*{Acknowledgement}
This research was supported by the Alkhawarizmi AI Project (grant number: Alkhawarizmi/2020/34).

\bibliographystyle{IEEEtran}
\bibliography{refs}

\begin{thebibliography}{10}
\providecommand{\url}[1]{#1}
\csname url@samestyle\endcsname
\providecommand{\newblock}{\relax}
\providecommand{\bibinfo}[2]{#2}
\providecommand{\BIBentrySTDinterwordspacing}{\spaceskip=0pt\relax}
\providecommand{\BIBentryALTinterwordstretchfactor}{4}
\providecommand{\BIBentryALTinterwordspacing}{\spaceskip=\fontdimen2\font plus
\BIBentryALTinterwordstretchfactor\fontdimen3\font minus
  \fontdimen4\font\relax}
\providecommand{\BIBforeignlanguage}[2]{{%
\expandafter\ifx\csname l@#1\endcsname\relax
\typeout{** WARNING: IEEEtran.bst: No hyphenation pattern has been}%
\typeout{** loaded for the language `#1'. Using the pattern for}%
\typeout{** the default language instead.}%
\else
\language=\csname l@#1\endcsname
\fi
#2}}
\providecommand{\BIBdecl}{\relax}
\BIBdecl

\bibitem{hullermeier2019}
\BIBentryALTinterwordspacing
E.~H{\"u}llermeier and W.~Waegeman, ``Aleatoric and epistemic uncertainty in
  machine learning: an introduction to concepts and methods,'' \emph{Machine
  Learning}, 2019. [Online]. Available:
  \url{https://link.springer.com/content/pdf/10.1007/s10994-021-05946-3.pdf}
\BIBentrySTDinterwordspacing

\bibitem{sideygibbons2019}
\BIBentryALTinterwordspacing
J.~A. Sidey-Gibbons and C.~J. Sidey-Gibbons, ``Machine learning in medicine: a
  practical introduction,'' \emph{BMC medical research methodology}, vol.~19,
  no.~1, pp. 1--8, 2019. [Online]. Available:
  \url{https://bmcmedresmethodol.biomedcentral.com/track/pdf/10.1186/s12874-019-0681-4.pdf}
\BIBentrySTDinterwordspacing

\bibitem{niknam2019}
\BIBentryALTinterwordspacing
S.~Niknam, H.~S. Dhillon, and J.~H. Reed, ``Federated learning for wireless
  communications: Motivation, opportunities, and challenges,'' \emph{IEEE
  Communications Magazine}, 2019. [Online]. Available:
  \url{https://ieeexplore.ieee.org/ielx7/35/8795161/08815113.pdf}
\BIBentrySTDinterwordspacing

\bibitem{hanjri2023federated}
M.~E. Hanjri, H.~Kabbaj, A.~Kobbane, and A.~Abouaomar, ``Federated learning for
  water consumption forecasting in smart cities,'' \emph{arXiv preprint
  arXiv:2301.13036}, 2023.

\bibitem{el2023new}
M.~El~Hanjri, A.~Abouaomar, and A.~Kobbane, ``New architecture conception for
  water distribution network in smart home,'' in \emph{2023 International
  Wireless Communications and Mobile Computing (IWCMC)}.\hskip 1em plus 0.5em
  minus 0.4em\relax IEEE, 2023, pp. 1591--1596.

\bibitem{lyu2020}
\BIBentryALTinterwordspacing
L.~Lyu, H.~Yu, X.~Ma, L.~Sun, J.~Zhao, Q.~Yang, and P.~S. Yu, ``Privacy and
  robustness in federated learning: Attacks and defenses,'' \emph{arXiv
  preprint arXiv:2012.03191}, 2020. [Online]. Available:
  \url{https://arxiv.org/pdf/2012.03191.pdf}
\BIBentrySTDinterwordspacing

\bibitem{kabbajdistfl}
H.~Kabbaj, M.~El~Hanjri, A.~Kobbane, R.~El-Azouzi, and A.~Abouaomar, ``Distfl:
  An enhanced fl approach for non trusted setting in water distribution
  networks.''

\bibitem{bonawitz2017}
\BIBentryALTinterwordspacing
K.~Bonawitz, V.~Ivanov, B.~Kreuter, A.~Marcedone, H.~B. McMahan, S.~Patel,
  D.~Ramage, A.~Segal, and K.~Seth, ``Practical secure aggregation for
  privacy-preserving machine learning,'' \emph{Proceedings of the 2017 ACM
  SIGSAC Conference on Computer and Communications Security}, 2017. [Online].
  Available: \url{https://dl.acm.org/doi/pdf/10.1145/3133956.3133982}
\BIBentrySTDinterwordspacing

\bibitem{fedavgm}
\BIBentryALTinterwordspacing
T.~H. Hsu, H.~Qi, and M.~Brown, ``Measuring the effects of non-identical data
  distribution for federated visual classification,'' \emph{CoRR}, vol.
  abs/1909.06335, 2019. [Online]. Available:
  \url{http://arxiv.org/abs/1909.06335}
\BIBentrySTDinterwordspacing

\bibitem{fedprox}
\BIBentryALTinterwordspacing
A.~K. Sahu, T.~Li, M.~Sanjabi, M.~Zaheer, A.~Talwalkar, and V.~Smith, ``On the
  convergence of federated optimization in heterogeneous networks,''
  \emph{CoRR}, vol. abs/1812.06127, 2018. [Online]. Available:
  \url{http://arxiv.org/abs/1812.06127}
\BIBentrySTDinterwordspacing

\bibitem{scaffold}
\BIBentryALTinterwordspacing
S.~P. Karimireddy, S.~Kale, M.~Mohri, S.~J. Reddi, S.~U. Stich, and A.~T.
  Suresh, ``{SCAFFOLD:} stochastic controlled averaging for on-device federated
  learning,'' \emph{CoRR}, vol. abs/1910.06378, 2019. [Online]. Available:
  \url{http://arxiv.org/abs/1910.06378}
\BIBentrySTDinterwordspacing

\bibitem{moon}
\BIBentryALTinterwordspacing
Q.~Li, B.~He, and D.~Song, ``Model-contrastive federated learning,''
  \emph{CoRR}, vol. abs/2103.16257, 2021. [Online]. Available:
  \url{https://arxiv.org/abs/2103.16257}
\BIBentrySTDinterwordspacing

\bibitem{fedyn}
\BIBentryALTinterwordspacing
D.~A.~E. Acar, Y.~Zhao, R.~M. Navarro, M.~Mattina, P.~N. Whatmough, and
  V.~Saligrama, ``Federated learning based on dynamic regularization,''
  \emph{CoRR}, vol. abs/2111.04263, 2021. [Online]. Available:
  \url{https://arxiv.org/abs/2111.04263}
\BIBentrySTDinterwordspacing

\bibitem{fedgen}
\BIBentryALTinterwordspacing
Z.~Zhu, J.~Hong, and J.~Zhou, ``Data-free knowledge distillation for
  heterogeneous federated learning,'' \emph{CoRR}, vol. abs/2105.10056, 2021.
  [Online]. Available: \url{https://arxiv.org/abs/2105.10056}
\BIBentrySTDinterwordspacing

\bibitem{fedlc}
J.~Zhang, Z.~Li, B.~Li, J.~Xu, S.~Wu, S.~Ding, and C.~Wu, ``Federated learning
  with label distribution skew via logits calibration,'' 2022.

\bibitem{fedavg}
\BIBentryALTinterwordspacing
A.~Fallah, A.~Mokhtari, and A.~Ozdaglar, ``Personalized federated learning with
  theoretical guarantees: A model-agnostic meta-learning approach,'' in
  \emph{Advances in Neural Information Processing Systems}, H.~Larochelle,
  M.~Ranzato, R.~Hadsell, M.~Balcan, and H.~Lin, Eds., vol.~33.\hskip 1em plus
  0.5em minus 0.4em\relax Curran Associates, Inc., 2020, pp. 3557--3568.
  [Online]. Available:
  \url{https://proceedings.neurips.cc/paper_files/paper/2020/file/24389bfe4fe2eba8bf9aa9203a44cdad-Paper.pdf}
\BIBentrySTDinterwordspacing

\bibitem{maml}
\BIBentryALTinterwordspacing
C.~Finn, P.~Abbeel, and S.~Levine, ``Model-agnostic meta-learning for fast
  adaptation of deep networks,'' \emph{CoRR}, vol. abs/1703.03400, 2017.
  [Online]. Available: \url{http://arxiv.org/abs/1703.03400}
\BIBentrySTDinterwordspacing

\bibitem{dirichlet}
\BIBentryALTinterwordspacing
T.~H. Hsu, H.~Qi, and M.~Brown, ``Measuring the effects of non-identical data
  distribution for federated visual classification,'' \emph{CoRR}, vol.
  abs/1909.06335, 2019. [Online]. Available:
  \url{http://arxiv.org/abs/1909.06335}
\BIBentrySTDinterwordspacing

\end{thebibliography}

\end{document}